\documentclass{article} 
\usepackage{iclr2022_conference,times}

\usepackage{amsmath,amsfonts,bm}

\def\eqref#1{equation~\ref{#1}}

\def\1{\bm{1}}

\DeclareMathAlphabet{\mathsfit}{\encodingdefault}{\sfdefault}{m}{sl}
\SetMathAlphabet{\mathsfit}{bold}{\encodingdefault}{\sfdefault}{bx}{n}

\usepackage{hyperref}
\usepackage{url}
\usepackage{tabularx}
\usepackage{graphicx}
\usepackage{booktabs}
\usepackage{algpseudocode}
\usepackage{algorithm}
\usepackage{CJKutf8}
\usepackage{subcaption}

\algblock{Input}{EndInput}
\algnotext{EndInput}
\algblock{Output}{EndOutput}
\algnotext{EndOutput}
\newcommand{\Desc}[2]{\State \makebox[5em][l]{#1}#2}

\newcommand{\modelone}{Multi-modal Caption Proposal}
\newcommand{\acrone}{MCProp}
\newcommand{\modeltwo}{Caption Re-Rank}
\newcommand{\acrtwo}{CRank}

\title{Transformer-Based Multi-modal Proposal and Re-Rank for Wikipedia Image-Caption Matching}

\author{Nicola Messina \\
Inst. of Information Science and Technologies\\
National Research Council\\
Pisa, Italy\\
\texttt{\small nicola.messina@isti.cnr.it}\\
\And 
Davide Alessandro Coccomini \\ Inst. of Information Science and Technologies\\
National Research Council\\
Pisa, Italy\\
\texttt{\small davidealessandro.coccomini@isti.cnr.it}\\
\And 
Andrea Esuli  \\
Inst. of Information Science and Technologies\\
National Research Council\\
Pisa, Italy\\
\texttt{\small andrea.esuli@isti.cnr.it}
\And
Fabrizio Falchi \\
Inst. of Information Science and Technologies\\
National Research Council\\
Pisa, Italy\\
\texttt{\small fabrizio.falchi@isti.cnr.it} 
}

\iclrfinalcopy 
\begin{document}

\maketitle
\begin{abstract}
With the increased accessibility of web and online encyclopedias, the amount of data to manage is constantly increasing. In Wikipedia, for example, there are millions of pages written in multiple languages. These pages contain images that often lack the textual context, remaining conceptually floating and therefore harder to find and manage.
In this work, we present the system we designed for participating in the \textit{Wikipedia Image-Caption Matching} challenge on Kaggle, whose objective is to use data associated with images (URLs and visual data) to find the correct caption among a large pool of available ones. A system able to perform this task would improve the accessibility and completeness of multimedia content on large online encyclopedias. Specifically, we propose a cascade of two models, both powered by the recent Transformer model, able to efficiently and effectively infer a relevance score between the query image data and the captions. We verify through extensive experimentation that the proposed two-model approach is an effective way to handle a large pool of images and captions while maintaining bounded the overall computational complexity at inference time.
Our approach achieves remarkable results, obtaining a normalized Discounted Cumulative Gain (nDCG) value of 0.53 on the private leaderboard of the Kaggle challenge.
\end{abstract}

\section{Introduction}

In recent years, deep learning techniques have achieved outstanding results in many computer vision and language tasks, and numerous attempts have been made to merge the two worlds.
A big step forward was achieved with the introduction of Transformers \citep{vaswani2017attention} which proved to be extremely powerful both in the field of natural language processing with models like BERT \citep{devlin2019bert}, ELMo \citep{peters2018deep}, GPT-3 \citep{brown2020language}, RoBERTa \citep{liu2019roberta} and ALBERT \citep{lan2020albert}, and in the field of computer vision with the introduction of Vision Transformers \citep{dosovitskiy2021image} and its variants like Swin Transformer \citep{liu2019roberta}, CrossViT \citep{chen2021crossvit} and Twins-SVT \citep{chu2021twins}. Given that Transformers are so effective in both fields, they were used to extract common representations between multi-modal data so that they could later be compared or processed altogether. For example, \citet{akbari2021vatt} use Transformers to extract common representations between audio, video, and text. 

Among all the emerging multi-modal tasks, image-caption matching is becoming really important, especially for scalable and efficient cross-modal retrieval \citep{messina2021finegrained,messina2021transformer}. Image-caption matching involves associating an image with the text that best describes it. It can be used to find the most relevant images for a given query text (\textit{text-to-image retrieval}) or vice-versa (\textit{image-to-text retrieval}). These are important challenges that can make multimedia content more accessible and complete. Text-to-image retrieval has important applications in multimedia search engines, where a natural language phrase is used to search for visual content \citep{amato2021visione}.

In this paper, we focus on image-to-text retrieval. Specifically, we participated to the recent Kaggle competition organized by the WikiMedia foundation\footnote{\url{https://www.kaggle.com/c/wikipedia-image-caption/overview}} that concerns the retrieval of captions from Wikipedia pages associated with a certain image. Both textual and visual information (image URL and image itself) can be used to find the caption. This challenge is motivated by the fact that the majority of images on Wikipedia articles do not have any written context connected to the image. Current solutions rely on simple methods based on translations or page interlinks, which have limited coverage. As of now, even the most advanced computer vision image captioning cannot handle complex semantics. 


For these reasons, we propose a cascade of two image-text matching models based on large pre-trained Transformer models. The first model is based on the common space matching approach and uses XLM-RoBERTa and CLIP as text and image feature extractors respectively. Being very efficient at inference time, this model is used to propose candidates for the second model, a fine-tuned XLM-RoBERTa pairwise classifier, less efficient but more accurate, that outputs the final caption candidates for each image in the test set. Our approach achieves the fifth position on the final private leaderboard, with a nDCG of 0.53.

The code for reproducing our results is publicly available on GitHub\footnote{ \url{https://github.com/mesnico/Wiki-Image-Caption-Matching}}.

\section{Related Works}


Our objective is to solve a multi-modal matching task, where we use image information to retrieve textual captions. The captions are chosen among a wide set of candidates.
Sentence matching obtained a huge boost in the last few years, thanks to large pre-training of Transformer models, such as BERT \citep{devlin2019bert}, RoBERTa \citep{liu2019roberta}, and multi-lingual versions of them like XLM-RoBERTa \citep{conneau2020unsupervised}. These models have been recently extended to work with images. Some works use BERT-like processing on both visual and textual modalities, such as ViLBERT \citep{lu2019vilbert}, ImageBERT \citep{qi2020imagebert}, Pixel-BERT \citep{huang2020pixelbert}, or VL-BERT \citep{Su2020VL-BERT}. Nevertheless, all these methods require several network evaluations that scale quadratically with the number of items in the inference set. In fact, all the possible image-caption pairs should be input into the network to obtain the matching score.
For this reason, many methods rely instead on the projection of visual and textual information into the same common space, where only a simple dot-product is needed to obtain the similarity between a given pair.
In particular, \citet{vsepp2018faghri} use VGG and ResNets visual features extractors, together with an LSTM for sentence processing, and they match images and captions exploiting hard-negatives at training time. 
With their VSRN architecture, \citet{li2019} introduce a visual reasoning pipeline built of Graph Convolution Networks (GCNs) and a GRU to sequentially reason on the different image regions.
Differently, in \citet{sarafianos2019adversarial} an adversarial learning method is proposed, and a discriminator is used to learn modality-invariant representations. The method proposed by \citet{guo2020associating} consists of a contextual attention-based LSTM-RNN which can selectively attend to salient regions of an image at each time step;  they employ a recurrent canonical correlation analysis to find hidden semantic relationships between regions and words. In \cite{Vo_2019_CVPR} the authors proposed a system for combining image and text features for image retrieval. They introduced a fusion approach called Text Image Residual Gating (TIRG), in which the image feature is first gated and then added to a residual feature which works as a \textit{modification} feature. They defeat other image-text fusion techniques on the task of image retrieval given a base image and a modification text as query. 
\citet{wu2019learning} used triplet and angular loss to project the image and sentence features into the same common space, while \citet{qu2020context} use BERT as a language model and an adaptive gating self-attention module to obtain context-enhanced visual features, projecting them into the same common space using cosine similarity.
Transformer networks have been used in image-text matching in \citet{messina2021transformer,messina2021finegrained} for the task of multi-modal large-scale information retrieval. They introduced a novel disentangled transformer architecture that separately reasons on the two different modalities and enforces a final common abstract concept space.

Differently from all these works, in our efficient candidate proposal model we dealt with multi-modal queries composed of a text (the image URL) and the image itself. Therefore, we have a slightly different setup, in which an (image, text) pair is used to retrieve the captions (another text).

\section{Method}
The data provided in the Wikipedia Image-Caption Matching challenge (on Kaggle) consists of three main fields: \textit{image URLs}, \textit{images}, and \textit{captions}. The challenge consists of finding the most relevant caption given the image URLs, the images, or both as a query.
Given that the test set is composed of around $n_t=92K$ elements, using a large Transformer to compute relevance score for every (query, target) pair is infeasible, as we would need to compute $n_t^2$ relevance scores to get the ranking for the whole test set.
Driven by this concern, we decided to adopt a cascade of two different models to produce the final rankings. 
The first one, which we call \modelone\ (\acrone) model, employs both the textual information in the image URLs and the visual information in the images as a compound query to infer the caption. This model projects queries and captions in the same common space, where cosine similarity is used to measure the similarity between a query and a caption. With this model, efficient k-nearest neighbor search can be performed to create a rank for every query of all the $n_t$ captions. 
The top-ranked elements are then used as candidates by the second model, called \modeltwo\ (\acrtwo) model. This is a large Transformer fine-tuned for pair classification, i.e., a binary classifier that classifies a (query, caption) pair as either a match or a non-match.
This second model employs only the textual information in the image URL to infer the caption without relying again on the visual information.
The highest match probabilities returned by \acrone\ determine the top-5 captions selected for every image, as requested by the challenge.

The backbone for text processing is common to the two models. Specifically, we used the XLM-RoBERTa$_{Base}$ model \citep{conneau2020unsupervised}, which employs byte-level Binary Pair Encoding \citep{wang2020neural} with a vocabulary of 50k tokens, 12 encoder layers, a hidden state of size 768, and 8 attention heads, with a total of 125M parameters.
The pre-trained XLM-RoBERTa$_{Base}$ model is the result of training on documents in 100 languages from the CommonCrawl dataset \citep{wenzek2020}.

Following, we present in detail the two models.

\subsection{The \modelone\ model}
\label{sec:matching-model}
The core idea of the \modelone\ method is to represent the query and target data in a common feature space. In this space, we can calculate the similarity in an efficient and scalable manner using cosine similarities.
The model comprises two pipelines, the \textit{query} pipeline and the \textit{caption} pipeline. In turn, a query is composed of an \textit{image URL} and an \textit{image}. Therefore, in total, we need to process two textual fields and one visual field.

Visual features are extracted from images via the image encoder part of a CLIP network \citep{radford2021learning}. CLIP is a powerful multi-modal model composed of an image encoder and a text encoder that is trained to predict the correct visual-textual pairings. Being pre-trained on a multi-modal task, the image encoder module is a very good fit for our task.

We do not use the textual pipeline of CLIP as our textual backbone. The main reason for this is that this challenge is inherently multilingual, and CLIP is not trained on multiple languages. For this reason, we use instead a pre-trained large language Transformer model, XLM-RoBERTa \citep{conneau2020unsupervised}, as a textual pipeline. 

\begin{figure*}[t]
  \centering
  \includegraphics[page=1,width=\textwidth]{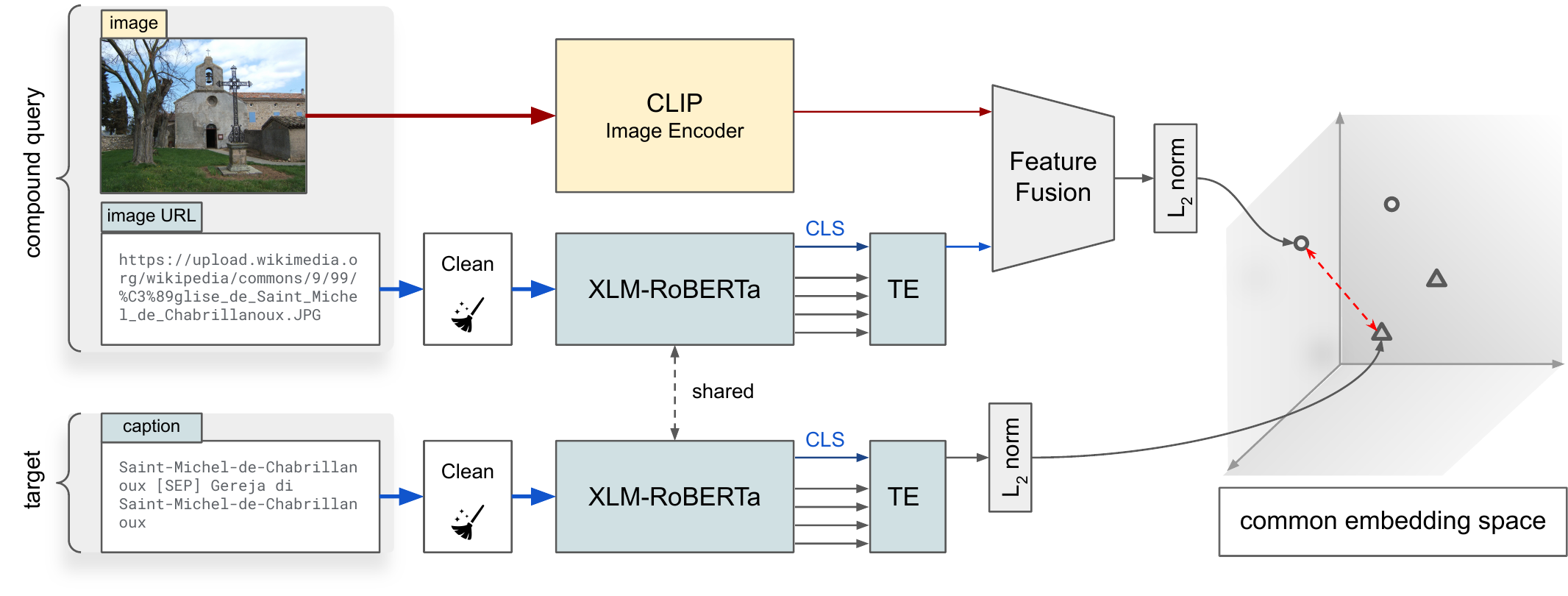}
  \caption{The \modelone\ model. The XLM-RoBERTa backbone is shared among the query and caption pipelines, and the representations are further specialized using downstream Transformer Encoders (TE).}
  \label{fig:common_feature_space}
\end{figure*}

The overall architecture is shown in Figure \ref{fig:common_feature_space}.
Specifically, the image encoder of CLIP outputs an aggregated visual feature $\bar{\mathbf{v}}$; differently, XLM-RoBERTa outputs a set $\{\mathbf{w}_1, \mathbf{w}_2,\ldots,\mathbf{w}_M\}$ of textual features, that are used by the heads of the original model for solving the many downstream tasks. Instead, we post-process these features by means of other Transformer Encoder layers in a way similar to TERN \citep{messina2021transformer} and TERAN \citep{messina2021transformer}, obtaining $\{\mathbf{w}'_1, \mathbf{w}'_2,\ldots,\mathbf{w}'_M\}$. We use the token embedding from the first element of the output sequence, the CLS token, as a final representation for the input text: $\mathbf{c} = \mathbf{w}'_1$.
Notice that the XLM-RoBERTa backbone is shared among the image URL and the caption pipelines. In order to specialize the representations to the downstream task, the two downstream Transformer Encoder modules from the two pipelines do not share the weights. Concerning input preprocessing, the image URL is cleaned by removing the extension and the part preceding the actual filename. 
All the underscores are also replaced by a space.

The image URL and the image are fused using an attentive fusion module described in the next paragraph.

\subsubsection{Attentive Feature Fusion}
Since two different modalities (the image URL and the image) are used as a compound query to infer the image caption, it would be useful to automatically infer the relative importance of the two components of the query to solve the matching task.
The \textit{attentive feature fusion} module, inspired by other works in this direction \citep{hazarika2018self,hori2017attention}, serves precisely this purpose. 
This module is composed of a sub-network that computes two attention values, one for each query component. Specifically, the network is a simple MLP with a final sigmoid layer that takes as input the concatenation among the two input vectors $\mathbf{u}$ (image URL) and $\mathbf{v}$ (the image) and outputs two scalars, $\alpha_{u}$ and $\alpha_{v}$:
\begin{align}
    \alpha_u, \alpha_v = \text{sigmoid}(\text{MLP}([\mathbf{u}, \mathbf{v}]))
\end{align}
where $[\cdot, \cdot]$ is concatenation.
Thanks to the final sigmoid layer, these values lay in the range [0, 1].
Those values are then used for computing the final query representation $\mathbf{q}$ as a weighted average between the normalized input vectors:
\begin{align}
    \mathbf{q} = \alpha_u \frac{\mathbf{u}}{||\mathbf{u}||} + \alpha_v \frac{\mathbf{v}}{||\mathbf{v}||}
\end{align}
The vectors are normalized so their intrinsic magnitude is 1. Doing so, the $\alpha_{u}$ and $\alpha_{v}$ values are forced to be directly comparable and more easily interpretable.

\subsubsection{Training}
In order to match images and captions in the same common space, we use a hinge-based triplet ranking loss, focusing the attention on hard negatives, as in \citet{vsepp2018faghri,li2019,messina2021finegrained,messina2021transformer}.
Specifically, given the final query representation $\mathbf{q}$ and the target caption feature $\mathbf{c}$ we use the following loss function:
\begin{equation}
\begin{split}
    L_{match}(\mathbf{q}, \mathbf{c}) = \max_{\mathbf{c}'} [\alpha + S(\mathbf{q}, \mathbf{c}') - S(\mathbf{q}, \mathbf{c})]_+ + \max_{\mathbf{q}'} [\alpha + S(\mathbf{q}', \mathbf{c}) - S(\mathbf{q}, \mathbf{c})]_+,
\end{split}
\end{equation}
where $[x]_+ \equiv \max(0, x)$.
$S(\mathbf{q},\mathbf{c})$ is the similarity function between the query vector and the target caption features. We used the standard cosine similarity as $S(\cdot, \cdot)$.
As in \citet{vsepp2018faghri}, the hard negatives are sampled from the mini-batch and not globally, for performance reasons.


\subsection{The \modeltwo\ model}
\label{sec:modeltwo}

\begin{figure}[t]
  \centering
  \includegraphics[page=2,width=.6\linewidth]{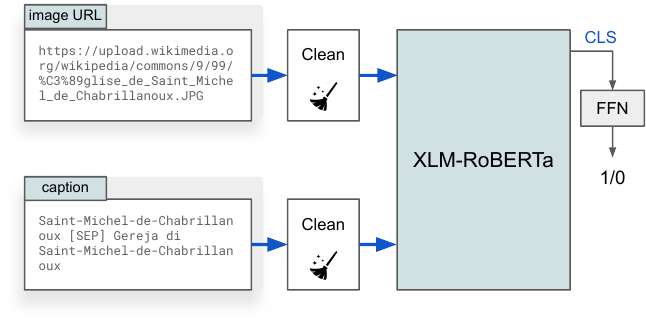}
  \caption{The \modeltwo\ model. It uses an XLM-RoBERTa masked language model pre-trained on the CommonCrawl dataset and fine-tuned on the image URL-caption match classification task.}
  \label{fig:modeltwo}
\end{figure}

The \modeltwo\ (\acrtwo) model is a binary classifier based on the XLM-RoBERTa model.
More specifically, the network consists of the XLM-RoBERTa model, i.e., the encoder part of a Transformer model, with the pooled output of the CLS token connected to a linear layer with an output size equal to the number of labels. The overall architecture is shown in Figure \ref{fig:modeltwo}.

The classification task aims to determine if an image URL and a caption match or not.
We use a processed version of the image URL to represent the image. 
We do not use visual information from the image in this phase.
As for the \modelone\ model, the URL is processed by removing any URL of the path component preceding the actual filename and any file type extension. 
Any underscore character in the remaining string is replaced by a space.
The input of the matching process is the concatenation of the processed URL and the caption text, with a SEP token separating them.

\subsubsection{Training}

To fine-tune the pre-trained model for our classification task, we trained \acrtwo\ using all the (image URL, caption) pairs available in the training dataset.
The dataset explicitly defines only examples of matches. To get examples of non-matching pairs, we used a simple negative sampling strategy that randomly pairs image URLs and captions from the dataset.
We generated several non-matching pairs equal to the matching ones, obtaining a training set of about 74 million examples.
The training process used a batch size of 64, with each batch containing an equal amount of matching and non-matching pairs.
We trained it for 2 epochs, using the Adam algorithm with weight decay \citep{loshchilov2017decoupled}, requiring 65 hours per epoch on a single NVidia RTX2080 GPU.

\subsubsection{Selection of candidates for classification}

The number of classifications required for the image-caption match task we are facing grows with the square of the size of the test set.
Each classification requires passing the string representing the (image URL, caption) pair through the XLMRoBERTa model, which takes a non-negligible time.
This is structurally different from the \acrone\ model, in which images and caption are projected in the common space separately, thus with a linear cost with respect to the test set size.
For the \acrone\ model, the quadratic cost is limited to the computation of the cosine similarities between the resulting vectors, which is a faster operation that can also be computed in parallel much easily.

The overall cost for \acrtwo\ can rapidly pass the limit of available computational and time resources.
With a single NVidia RTX2080 GPU the time required to compute the classification scores for all the (image URL, caption) pairs in the test set is more than three months.
For comparison, computing all the pairwise similarities on embedding vectors of length 768 for the same test set requires eight minutes on a desktop CPU.
Multiple GPUs can reduce the cost to a more manageable time, yet the quadratic nature of the process still make it not scalable to larger dataset.
For this reason, we employed this two-steps approach, in which a faster method, e.g., \acrone, selects a smaller set of promising candidates to be processed by \acrtwo.
Having a fixed number of candidate captions for image makes the cost linear with respect to the dataset size.
Using 1,000 candidate captions per image, determined using \acrone\ or other methods, it takes 27 hours to apply \acrtwo\ to the 92K image URLs from the test set.

\section{Experiments}

\subsection{Dataset}
The dataset used for our experiments is the one publicly released on the Kaggle competition page. 
It is based on Wikipedia-based Image Text Dataset (WIT) \citep{srinivasan2021wit} and contains 37.6 million entity-rich image-text examples with 11.5 million unique images across 108 Wikipedia languages.
There are at least 12,000 examples for each language, making the dataset particularly interesting for building a model that is not necessarily relegated to a specific language. Each example contains an image URL, from which the image can be downloaded, and the target caption. 

The dataset is already divided into a training and a test set. The training set contains for each (image, image URL) pair the associated caption describing the image. On the other hand, the test set separately comprises a list of (image, image URL) pairs that compose the query and a list of captions not paired with the given queries. Each of these two lists contains 92,367 entries.
The ground-truth for the test set, i.e., the (image, caption) pairs, is not released. 
The only way to obtain the results on the test set is to submit the inferred top-5 captions for each query to the Kaggle evaluation server. 

\subsection{Evaluation}
The quality of the obtained ranking is calculated considering the top 5 most similar captions for each image and applying the normalized Discounted Cumulative Gain (nDCG), a variant of the classical Discounted Cumulative Gain (DCG). The reasoning behind DCG is to penalize relevant items that are preceded by non-relevant items in the ordered list of results. The DCG grows with the exponential of the graded relevance of an item, while it is inversely proportional to the logarithm of the item rank. The DCG at a particular rank position $p$ is defined as follows:
$$DCG_p = \sum_{i=1}^p \frac{2^{\text{rel}_i} - 1}{\log_2{(i+1)}}, $$
where $\text{rel}_i$ is the graded relevance of the item at position $i$ in the results.
The nDCG normalizes DCG by its maximum theoretical value and thus returns values in the $[0,1]$ range.
%
%

Besides the nDCG metric employed in the challenge, we also compute the recall@K metric on our validation sets. This metric is widely-used when there is only one relevant item for every given query, like in this case. The recall@K measures the percentage of queries able to retrieve the correct result among the first $K$ retrieved items.



\subsection{Preliminary Results on the \acrone\ model}

\newcolumntype{C}{>{\centering\arraybackslash}X}
\newcolumntype{R}{D{,}{\pm}{1.6}}
\newcolumntype{L}{>{\raggedright\arraybackslash}p{4cm}}
\begin{table}[t]
\caption{Caption retrieval results for the \modelone\ model on the validation set.}
\label{tab:model-1-results}
\begin{center}
\begin{tabular}{Lcccc}
\toprule
& \multicolumn{3}{c}{\textbf{Recall@K}} & \textbf{nDCG$_5$}\\
\cmidrule(lr){2-4}
\textbf{Model} & \multicolumn{1}{c}{K=1} & \multicolumn{1}{c}{K=5} & \multicolumn{1}{c}{K=10} & \\
\midrule
\acrone\ no-imgs & 48.1 & 55.9 & 58.5 & 0.522 \\
\acrone\ imgs & 48.0 & 57.9 & 62.2 & 0.533 \\
\acrone\ imgs, att-fusion & \textbf{48.4} & \textbf{58.6} & \textbf{62.7} & \textbf{0.538} \\
\bottomrule
\end{tabular}
\end{center}
\end{table}

For training the \modelone\ model, we reserved 10,000 examples chosen randomly from the given training set for validating.
For validating the model, we used the main metric used for the challenge, the nDCG$_5$. However, we report also the Recall@K, as it is one of the main metrics used in cross-modal retrieval literature. 

We used CLIP provided with a ViT as the backbone for the visual pipeline. Instead, the language backbone is a XLM-RoBERTa, a large Transformer Encoder model pre-trained on a large and multilingual textual corpus. Both the visual and textual backbones were frozen during the training.

As a query, we tried two different configurations. In a first experiment, we used a compound query built by employing both the image URL (processed with the textual pipeline) and the image (processed with the image pipeline). In the second configuration, instead, we used only the image URL. These two experiments aim to understand the role of images in solving the matching task, given that the image URL alone seems already sufficient in most cases.
When both the image URL and the image are used as a query, as mentioned in \ref{sec:matching-model}, we used two different fusion techniques: straightforward concatenation or the attentive fusion mechanism.

\begin{figure}[t]
  \centering
  \includegraphics[width=.7\linewidth]{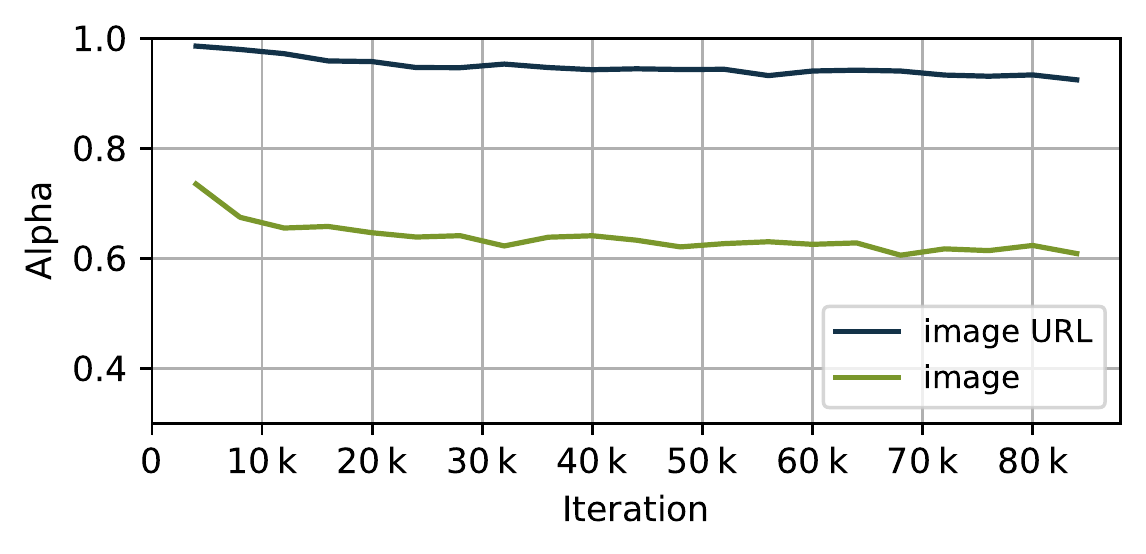}
  \caption{Weights estimated by the model for the image URL and the image ($\alpha_{u}$ and $\alpha_{v}$) on the full 10K validation set, as training progresses.}
  \label{fig:alphas}
\end{figure}

\paragraph{Discussion}
Table \ref{tab:model-1-results} shows the results reached on the validation set for the different experiments. In particular, we can see that images slightly increase the overall metrics. Nevertheless, the performance increase is not substantial, especially for recall@1, which seems to downgrade. 
Better results are obtained when using the attentive fusion approach to merge image URLs with visual information.
The use of the attentive fusion allowed us to inspect the relevance given by the model to the two different modalities that compose the query. Figure \ref{fig:alphas} shows the evolution of the two attention values on the full 10K validation set. Notice that the fusion module does not output normalized weights (their sum is not 1), as it is left free to adjust the scale of the input features without constraining their combination to be convex. As we can notice, the weight provided by the model for the visual pipeline remains at 65\% with respect to the weight assigned to the image URL pipeline. This confirms the evidence that the visual information contributes less to the matching task in this scenario.
In Appendix \ref{sec:bijective}, we explored another inference methodology that exploited the fact that captions assigned to an image are no more eligible for another image.



\subsection{Preliminary Results on the \acrtwo\ model}

The \acrtwo\ model takes as input a (image URL, caption) pair and returns a classification score.
The higher the score, the higher the confidence of a match.
Given a pool of candidate captions for an image URL, all the (image URL, caption) pairs must be classified by \acrtwo, sorting then the pairs by classification score, higher to lower.

As detailed in Section \ref{sec:modeltwo}, the computational complexity and cost of \acrtwo\ made this method not directly applicable to the test set without having access to substantial computational resources, which was not our case.
In order to test the performance of \acrtwo\ independently of the candidate selection method, we used a smaller validation set of 1,000 elements (held out from the training data).
The small size of this validation set made it possible to apply \acrtwo\ to the whole set without performing a candidate selection first.
We then applied several methods that act as candidate selectors.
Each candidate selector is used to rank all the captions, and then only the top-ranked 20\% of the captions are re-ranked using the classification scores produced by \acrtwo.
In this way, we can measure which method for selecting candidates works better in combination with \acrtwo.

\paragraph{Discussion}
Table \ref{tab:model-2-results} shows that \acrtwo\ obtains very high nDCG when applied to the whole validation set, placing the right caption in the right position 74\% of the time and 86\% of the time among the first 10 results.
As comparative baselines, we tested three methods.
We selected the top 5 most similar captions using the Levenshtein similarity with the cleaned image URL.
Also, we used the embedding vector for the CLS token produced by the XLM-RoBERTa model before the classification layer for each caption and each image URL to compute pairwise cosine similarity as the measure to select the top 5 captions.
Finally, we applied \acrone\ to the small validation set.
Comparing the baselines, it is interesting to see that the XLM-RoBERTa similarity method performs worse than the others. 
This indicates that the embeddings extracted from XLM-RoBERTa are so specialized for the classification task that they are no longer suitable for language representation.
The Levenshtein-based model obtains an average performance. This is a remarkable result considering that this method is fundamentally not able to handle pairs of texts in different languages.
On this smaller validation set, the \acrone method performs very well, placing the proper caption in the first position 57\% of the time.

We then used the baseline methods as a way to select a reduced set of candidates (20\% of the 1,000 elements in the validation set), which are then re-ranked by \acrtwo.
This two-steps procedure makes \acrtwo\ applicable to larger test sets.
In all cases, the \acrtwo\ re-rank produces a sensible improvement over the baselines.
When Levenshtein and XLM-RoBERTa are used the final scores are lower than the maximum achieved by \acrtwo\ on the full validation set.
This indicates that both Levenshtein and XLM-RoBERTa fail to put the correct result in the top 20\% of their ranks for a sensible number of cases.
On the contrary, the combination of \acrone\ and \acrtwo\ produces a slight increase in recall@1 and nDCG, with respect to the use of \acrtwo\ only.
This is due to a positive interaction between the two methods: (i) all the elements placed by \acrtwo\ in its top 5 positions are also placed by \acrone\ in its top 20\% positions; (ii) a few cases of tie in \acrtwo\ that caused the top result to be not the correct one are solved by \acrone\, that puts the tied element in the correct order\footnote{Whenever there is a tie in the \acrtwo\ ranking, the order of the tied elements in the list of candidate elements is preserved.}.

\newcolumntype{L}{>{\raggedright\arraybackslash}p{6.5cm}}
\begin{table}[t]
\caption{Caption retrieval results for the \modeltwo\ model, and candidate selector methods, and their combination, on the small validation set (1,000 elements).}
\label{tab:model-2-results}
\begin{center}
\begin{tabular}{Lcccc}
\toprule
& \multicolumn{3}{c}{\textbf{Recall@K}} & \textbf{nDCG$_5$} \\
\cmidrule(lr){2-4}
\textbf{Method} & \multicolumn{1}{c}{K=1} & \multicolumn{1}{c}{K=5} & \multicolumn{1}{c}{K=10}  &\\
\midrule
Levenshtein similarity & 32.2 & 38.6 & 40.8 & 0.356 \\
XLM-RoBERTa similarity & 17.6 & 24.5 & 27.5 & 0.213 \\
\acrone\ imgs, att-fusion & 57.7 & 70.9 & 76.2 & 0.647\\
\midrule
\acrtwo\ (100\%) & 74.2 & \textbf{82.7} & \textbf{86.4} & 0.789 \\
\midrule
\acrtwo\ on Levenshtein similarity (20\%) & 52.6 & 56.5 & 57.3 & 0.548\\
\acrtwo\ on XLM-RoBERTa similarity (20\%) & 48.8 & 53.2 & 53.7 & 0.512 \\
\acrtwo\ on \acrone\ imgs, att-fusion (20\%) &  \textbf{74.5} & \textbf{82.7} & \textbf{86.4} & \textbf{0.791} \\
\bottomrule
\end{tabular}
\end{center}
\end{table}

\subsection{Final Results}
This section reports the final results obtained on the private leaderboard of the Wikipedia challenge on Kaggle. Until the end of the competition, only the results on the public test set, composed of only 25\% of the full test set, were shown to the teams. Five submissions per day were accepted on the public leaderboard. Therefore, teams had the opportunity to optimize their methods on the small test subset. The results in the private leaderboard were instead computed at the end of the challenge, on 85\% of the test set. For all these reasons, despite not being publicly accessible, the results in the private leaderboard are more representative of the final model ranking.


In Table \ref{tab:test-results}, we report our models' performance on the private test set. We can notice how the choices made using the validation sets are well reflected also on the wider test corpus. In particular, the two-cascaded model pipeline obtains higher results among all the baseline methods. This confirms the ability of the \acrone\ model to propose relevant candidates and the proficiency of the \acrtwo\ to move correct items towards the top of the ranked list.
In the Appendix \ref{sec:translating}, we further tried to eliminate the need to explicitly deal with multiple languages by translating the whole test set into English. This trial did not improve the results reported in Table \ref{tab:test-results}, confirming the strength of multilingual models in solving this complex matching task.

Finally, in Table \ref{tab:leaderboard}, we report the results on the private leaderboard for the top-10 methods. More than 100 teams took part in this challenge. 
We positioned fifth in the competition.
The first participant significantly overperformed the rest of the participants, and also the second performed distinctly well.
Once the methods adopted by other participants will be published, it will be interesting to compare more aspects of the proposed solutions, including their computational complexity, their scalability, and the computational resources used to produce the submissions.

\newcolumntype{L}{>{\raggedright\arraybackslash}p{3.8cm}}
\begin{table}
\begin{subtable}[t]{0.48\textwidth}
\centering
\caption{nDCG$_5$ of our models on the test set. The \textit{Base} column reports results obtained using the given similarity model only; the \textit{\acrtwo\ } column shows the results using these methods to propose candidates for the \acrtwo\ model (1,000 candidates).}
\begin{tabular}{Lcc}
\toprule
\textbf{Similarity model} & \textbf{Base} & \textbf{\acrtwo} \\
\midrule
Levenshtein & 0.215 & 0.374\\
XLM-RoBERTa similarity & 0.175 & 0.269 \\
\acrone\ imgs, att-fusion & \textbf{0.421} & \textbf{0.533} \\
\bottomrule
\end{tabular}
\label{tab:test-results}
\end{subtable}
\hspace{4mm}
\begin{subtable}[t]{0.48\textwidth}
\centering
\caption{nDCG$_5$ of the top-10 performing methods on the private leaderboard.}
\begin{tabular}{cLc}
\toprule
\textbf{\#} & \textbf{Team Name} & \textbf{Score} \\
\midrule
1 & \begin{CJK*}{UTF8}{gbsn} 新东方人工智能研究院 \end{CJK*} & 0.734 \\
2 & sigs21group & 0.614 \\
3 & Basic Go & 0.559 \\
4 & Sadidul Islam & 0.535 \\
5 & \textbf{Fan Dani (ours)} & 0.533 \\
6 & GMago123 & 0.492 \\
7 & OddAsparagus11 & 0.444 \\
8 & Tetsuro Asano & 0.436 \\
9 & Lab\_EKB & 0.432 \\
10 & b0w1d & 0.414 \\
\bottomrule
\end{tabular}
\label{tab:leaderboard}
\end{subtable}
\caption{Final results on the private leaderboard (85\% test data).}
\end{table}

\section{Conclusions}
In this paper, we proposed a system able to match images with corresponding multilingual captions. This is an important tool for managing large encyclopedia websites, like Wikipedia, where most article images do not have any written context connected to the image. Driven by the power of recent Transformer-based models, we addressed the matching task using a cascade of two models. The first model, which we called \modelone\ model, uses CLIP image embeddings and XLM-RoBERTa word embeddings to project image information and captions into the same common space, where fast nearest-neighbor search can be performed. This model uses both image URL and visual information as a query, merging these data using an attentive feature fusion technique. The top candidates are then re-ranked using the second model, called \modeltwo\ model, which relies on the large pre-trained XLM-RoBERTa Transformer model fine-tuned on the Wikipedia training corpus. The cascaded pipeline is necessary, as the inference times for the \acrtwo\ model scale quadratically with the number of elements in the test set, and hardware limitations forced us to find a powerful upstream candidate proposal method. The results obtained on the validation sets show that the \modelone\ model is an effective model for candidate proposal, and \modeltwo\ is able to bring the correct results (chosen among the candidates) towards the top of the ranked list.

We participated in the Wikipedia  Image/Caption Matching challenge proposed on Kaggle, reaching the fifth position on the private leaderboard among more than 100 participating teams, with a system running on limited computational resources. 
Having a single performance measure allows us to make only a qualitative comparison. 
The publication of the approaches adopted by the participants will enable to have a quantitative comparison, taking into account also the complexity and resources required by each method to produce its results.

In the near future, we plan to leverage visual information also for the \acrtwo\ model and to perform feature extraction using different backbones, possibly pre-trained with diverse data.



\bibliography{biblio}
\bibliographystyle{iclr2022_conference}

\appendix
\section{Bijective matching}
\label{sec:bijective}
This challenge has an objective slightly different from the one underlying search systems. While in search systems two different queries possibly return the same relevant items, here we would like to obtain a bijective matching between the queries and the captions. In fact, every image seems to have its own description that cannot be shared with others.

We try to tackle this task at inference time by searching for the best assignment between queries and captions. We perform this association through a \textit{linear sum assignment} on the inferred score matrix. Formally, we are given the score matrix $\mathbf{S} \in \mathbb{R}^{n\times n}$. This matrix has $n$ rows, one for each query, and contains in its $i$-th row the scores computed between the $i$-th query and all the $n$ captions. The $n$ pairs of query and caption indexes $(q, c)$ that guarantee the best score assignment are obtained by solving the following linear optimization task:
\begin{align}
    \max & \sum_{i=1}^n \sum_{j=1}^n s_{i,j} x_{i,j} \\
    \text{s.t.} & \sum_{j=1}^n x_{i,j} = 1 \quad (i = 1, 2,\dotso, n),\\
    & \sum_{i=1}^n x_{i,j} = 1 \quad (j = 1, 2,\dotso, n), \\
    & x_{i, j} \in \{0, 1\} \quad (i, j = 1, 2,\dotso, n).
\end{align}
$\mathbf{X}$ is a permutation matrix found by the optimization algorithm such that $X_{q, c} = 1$.

We need the top-$k$ candidates for each query, and in particular, we require the top-5 for submitting to the challenge. However, a single run of the linear assignment procedure gives us only the top-1 best assignments for each query. To obtain the top-$k$, we run the linear assignment algorithm $k$ times, every time eliminating from the candidate pool the elements already assigned in the previous round. In this way, the elements assigned at the $i$-th run are no more eligible at the ($i+1$)-th run. We report the complete bijective matching procedure in Algorithm \ref{alg:bijective-matching}.

\begin{algorithm}[hbt!]
\caption{Bijective Matching}\label{alg:bijective-matching}
\begin{algorithmic}
\Input
\Desc{$\mathbf{S} \in \mathbb{R}^{n\times n}$}{Matrix of scores}
\Desc{$k$}{Number of top candidates for each query}
\EndInput
\Output
\Desc{$\mathbf{I} \in \mathbb{N}^{n\times k}$}{Indexes of top-k elements for each query}
\EndOutput

\For{$t \gets 1$ to $k$}
    \State $(\mathbf{q}, \mathbf{c}) \gets $\Call{LinearSumAssignment}{$\mathbf{S}$}
    \State $\mathbf{S}[\mathbf{q}, \mathbf{c}] \gets 0$ \Comment{Zero out the scores of assigned elements} 
    \State $\mathbf{I}[\mathbf{q}, t] = \mathbf{c}$ \Comment{Caption indexes are appended to the result}
\EndFor
\end{algorithmic}
\end{algorithm}

\newcolumntype{C}{>{\centering\arraybackslash}X}
\newcolumntype{R}{D{,}{\pm}{1.6}}
\newcolumntype{L}{>{\raggedright\arraybackslash}p{5cm}}
\begin{table}[t]
\caption{nDCG$_5$ on the \modelone\ model, with and without the bijective matching on both validation and test sets.}
\label{tab:bijective-matching}
\begin{center}
\begin{tabular}{Lcc}
\toprule
\textbf{Model} & \textbf{Validation} & \textbf{Test} \\
\midrule
\acrone\ imgs, att-fusion & 0.538 & 0.421 \\
\acrone\ imgs, att-fusion, bijective & 0.556 & 0.419 \\
\bottomrule
\end{tabular}
\end{center}
\end{table}

The final results for the bijective matching are shown in Table \ref{tab:bijective-matching}. We used the scores from the \modelone\ model computed on the 10K validation set. The bijective matching methodology seems to obtain the best results on the validation set. Unfortunately, it did not enhance the results obtained on the test set. This could be due to the unbalanced distribution of the output scores, which makes the whole model less and less linear as the size of the inference set increases. 

\section{Translating to a Common Language}
\label{sec:translating}
The (image URL, caption) pairs processed by the \acrtwo\ model may be formed by two pieces of text in different languages.
Our hypothesis in the definition of \acrtwo\ was that the multilingual XLM-RoBERTa model, pre-trained on a large multilingual corpus, would be able to classify the match/non-match between image URLs and captions independently of their language.

We also explored an alternative approach in which the captions and pieces of text resulting from cleaning the image URLs (see Section \ref{sec:modeltwo}) are all translated to English before being fed to \acrtwo.
The hypothesis supporting this approach is that pairwise translation models should be accurate in projecting all the text into a single language. The match/non-match classification task should be easier when run on a single language.
We extended this hypothesis to \acrone, using English-translated text instead of the original multilingual one.

Table \ref{tab:trans} reports the results from submissions that use the translation-based approach compared to our reference multilingual approach.
The translation impact is negative for both models, so we rejected the hypothesis that translation could benefit the task. 
Results support the idea of learning to project entities from different sources, either media or language, into a common space without resorting to performing any explicit and more complex translation.

\newcolumntype{L}{>{\raggedright\arraybackslash}p{4cm}}
\begin{table}[t]
\caption{Comparison of nDCG$_5$ of \acrone\ and \acrtwo\ using the original multilingual text or the English-translated version. Evaluated on submissions on the private leaderboard.}
\label{tab:trans}
\begin{center}
\begin{tabular}{Lcc}
\toprule
\textbf{Model} & \textbf{Multilingual} & \textbf{Translated} \\
\midrule
\acrone\ imgs, att-fusion & 0.421 & 0.395 \\
\acrtwo\ on \acrone\ imgs, att-fusion (1K candidates) & 0.533 & 0.499 \\
\bottomrule
\end{tabular}
\end{center}
\end{table}

\end{document}